\documentclass{svproc}
\usepackage{graphicx}
\usepackage{multicol}
\usepackage[bottom]{footmisc}

\usepackage[hidelinks]{hyperref}
\usepackage{placeins}
\usepackage{hhline}
\usepackage{booktabs}                                               % good-looking tables 
\usepackage{multirow}                                               % for tables
\usepackage{hyphenat}
\usepackage{subcaption}
\usepackage{amsmath}
\usepackage{url}

\usepackage{array}
\usepackage{adjustbox} 
\usepackage{caption}
\usepackage{float}       % For [H] placement
\usepackage{placeins}    % For \FloatBarrier

\begin{document}
\mainmatter              % start of a contribution
\title{Integrating SAINT with Tree-Based Models: A Case Study in Employee Attrition Prediction}
\titlerunning{Integrating SAINT with Tree-Based Models}  % abbreviated title (for running head)
\author{Adil Derrazi \and Javad Pourmostafa Roshan Sharami}
\authorrunning{A. Derrazi and J. Pourmostafa} % abbreviated author list (for running head)
%\tocauthor{First Author, Second Author, Third Author and Fourth Author}
%
\institute{CSAI Department, Tilburg University, The Netherlands\\
\email{derraziadil@gmail.com, J.Pourmostafa@tilburguniversity.edu}
}
%\institute{Princeton University, Princeton NJ 08544, USA,\\
%\email{lncs@springer.com}
%\and
%Universit\'{e} de Paris-Sud,
%Laboratoire d'Analyse Num\'{e}rique, B\^{a}timent 425,\\
%F-91405 Orsay Cedex, France}

\maketitle              % typeset the title of the contribution

\begin{abstract}
Employee attrition presents a major challenge for organizations, increasing costs and reducing productivity. Predicting attrition accurately enables proactive retention strategies, but existing machine learning models often struggle to capture complex feature interactions in tabular HR datasets. While tree-based models such as XGBoost and LightGBM perform well on structured data, traditional encoding techniques like one-hot encoding can introduce sparsity and fail to preserve semantic relationships between categorical features.

This study explores a hybrid approach by integrating SAINT (Self-Attention and Intersample Attention Transformer)-generated embeddings with tree-based models to enhance employee attrition prediction. SAINT leverages self-attention mechanisms to model intricate feature interactions. In this study, we explore SAINT both as a standalone classifier and as a feature extractor for tree-based models. We evaluate the performance, generalizability, and interpretability of standalone models (SAINT, XGBoost, LightGBM) and hybrid models that combine SAINT embeddings with tree-based classifiers.

Experimental results show that standalone tree-based models outperform both the standalone SAINT model and the hybrid approaches in predictive accuracy and generalization. Contrary to expectations, the hybrid models did not improve performance. One possible explanation is that tree-based models struggle to utilize dense, high-dimensional embeddings effectively. Additionally, the hybrid approach significantly reduced interpretability, making model decisions harder to explain. These findings suggest that transformer-based embeddings, while capturing feature relationships, do not necessarily enhance tree-based classifiers. Future research should explore alternative fusion strategies for integrating deep learning with structured data.

\end{abstract}

\section{Introduction} 
\label{sec:introduction}

This section introduces the research problem, motivation, and significance of predicting employee attrition. It highlights workforce management challenges, the limitations of traditional machine learning models for tabular data, and the potential of transformer-based approaches. The section also outlines the research questions, scientific contributions, and societal implications of the study.

\subsection{Problem Statement}
High employee turnover contributes to persistent labor shortages, as reflected in the 424,000 job vacancies reported in the Netherlands in 2023 \cite{cbs2024}. Organizations invest substantial resources in employee retention, yet high turnover persists due to competing job offers with better conditions \cite{awvn2024}.

Attrition has considerable financial and operational implications. Recruiting new employees is costly, and the training periods delay productivity. In industries requiring specialized skills, replacing personnel is even more challenging. Predicting employee attrition in advance enables human resources (HR) and management to implement retention strategies, reducing disruptions caused by unexpected departures.

HR datasets are structured as tabular data, where each row represents an employee, and each column corresponds to specific attributes such as job title, salary, or education level \cite{waters2018hr,cambridgespark2020hr}. However, predicting attrition from structured data presents unique challenges. Many machine learning models struggle to effectively capture interactions between categorical and numerical features, particularly when categorical features exhibit complex relationships. Addressing these challenges is critical for improving predictive accuracy in workforce analytics.

Tree-based models, such as \textit{XGBoost} and \textit{Random Forest}, are widely used for tabular data due to their high predictive performance \cite{chen2016xgboost}. However, they require categorical variables to be encoded, commonly through \textit{one-hot encoding}, which increases dimensionality and leads to potential overfitting and computational inefficiencies \cite{guo2016entity}. Additionally, one-hot encoding does not capture relationships between categories, such as similarities between job roles like ``Manager'' and ``Supervisor'' \cite{zheng2018feature}. Addressing these limitations is critical for improving predictive performance in employee attrition modeling.

\subsection{Proposed Approach}
Traditional encoding methods often fail to preserve complex relationships in categorical data. Transformer-based models such as \textit{SAINT} and \textit{TabTransformer} have demonstrated the ability to generate contextual embeddings that retain feature interactions \cite{huang2020tabtransformer,somepalli2021saint}. 

We propose a novel hybrid approach that combines transformer-based feature extraction with tree-based classification. Specifically, SAINT generates contextual embeddings from categorical features, which are then used as input for tree-based models such as XGBoost. Also, we evaluate SAINT as a standalone model to assess its individual predictive performance. We hypothesize that combining SAINT’s feature extraction with tree-based models will improve accuracy while maintaining interpretability.

\subsection{Research Question}
This study investigates the effectiveness of \textit{SAINT}-generated embeddings in improving tree-based models for employee attrition prediction. The main research question is:

\begin{itemize}
    \item \textit{What impact do SAINT-generated embeddings have on predictive performance compared to tree-based models using traditional encoding methods for employee attrition prediction?}
\end{itemize}

To address this question, the study compares the following models:

\begin{itemize}
    \item \textbf{Standalone models:}
    \begin{itemize}
        \item \textit{SAINT}: A transformer-based model designed for tabular data.
        \item \textit{XGBoost} and \textit{LightGBM}: Tree-based models optimized for structured data.
    \end{itemize}
    \item \textbf{Hybrid models:}
    \begin{itemize}
        \item \textit{SAINT-XGBoost}: \textit{SAINT}-generated embeddings used as input for \textit{XGBoost}.
        \item \textit{SAINT-LightGBM}: \textit{SAINT}-generated embeddings used as input for \textit{LightGBM}.
    \end{itemize}
\end{itemize}

\subsection{Scientific and Societal Relevance}
This research contributes to both scientific literature and practical workforce management by:
\begin{itemize}
    \item Exploring the use of transformer-based feature extraction for tabular data.
    \item Evaluating the impact of \textit{SAINT}-generated embeddings on predictive accuracy in tree-based models.
    \item Investigating trade-offs between model performance, generalizability, and interpretability.
\end{itemize}

From a societal perspective, improving attrition prediction helps organizations optimize workforce management, reduce hiring costs, and enhance employee retention strategies. Better understanding of attrition drivers enables targeted HR interventions, fostering ethical decision-making and a stable work environment, particularly in labor-constrained markets like the Netherlands.

\subsection{Key Findings}
\begin{itemize}
    \item \textbf{Predictive Performance:}  
    Tree-based models (\textit{XGBoost} and \textit{LightGBM}) outperformed all other models. Surprisingly, the hybrid pipelines (\textit{SAINT-XGBoost} and \textit{SAINT-LightGBM}) did not improve performance and, in some cases, performed worse than standalone SAINT. This suggests that SAINT-generated embeddings may not be well-aligned with tree-based classifiers, which are optimized for structured tabular inputs rather than high-dimensional feature representations.

    \item \textbf{Generalizability:}  
    The standalone tree-based models demonstrated strong generalization, with consistent cross-validation and test set performance. The \textit{SAINT} model also exhibited good generalizability, whereas the hybrid models showed performance degradation, indicating weaker generalization.

    \item \textbf{Interpretability:}  
    SHAP analysis (See Appendix~\ref{sec:shap_app}) showed that both tree-based models and \textit{SAINT} identified similar key predictors of attrition. However, computing SHAP values for \textit{SAINT} was significantly more resource-intensive, limiting its practical usability. Also, the interpretability of hybrid models was reduced, as embeddings obscured direct feature contributions.
\end{itemize}

\section{Related Work}

Employee attrition prediction has gained significant attention in human resource management due to its impact on organizational performance and the costs associated with employee turnover. Existing studies primarily focus on class imbalance handling and model performance comparisons, yet limited attention has been given to advanced encoding techniques for categorical variables in HR datasets \cite{app12136424,panigrahi2018employee,Mansor2021}. This section reviews traditional and modern approaches to predicting employee attrition, emphasizing advancements in handling categorical features.

\subsection{Traditional Approaches to Predicting Attrition}

Early models relied on statistical methods such as logistic regression to assess the relationship between employee characteristics and attrition \cite{10111843}. While widely used, these models struggle with complex and nonlinear patterns in HR datasets, limiting their predictive capabilities.

\subsection{Class Imbalance and Model Performance}

Recent research has shifted toward machine learning models capable of handling complex data structures. However, the predominant focus has been on addressing class imbalance rather than improving data representation. \cite{alduayj2018predicting} applied the Synthetic Minority Over-sampling Technique (SMOTE) to balance an imbalanced dataset before evaluating classifiers such as decision trees, random forests, and support vector machines. Their study emphasized the importance of class balancing but did not explore the role of advanced encoding methods.

\cite{panigrahi2018employee} compared logistic regression, decision trees, and ensemble methods on an imbalanced attrition dataset, highlighting the effects of class imbalance on predictive accuracy. Although feature selection was considered, the study primarily employed conventional preprocessing techniques without leveraging advanced encoding methods to capture feature relationships.

\subsection{Tree-Based Models and Categorical Data Representation}

Tree-based models, such as \textit{Random Forest} and \textit{Gradient Boosted Decision Trees (GBDT)}, are widely used for structured data due to their ability to model complex interactions \cite{chen2016xgboost}. \cite{alduayj2018predicting} found that tree-based models outperformed other classifiers in attrition prediction.

Despite their advantages, tree-based models require categorical variables to be converted into numerical representations, typically through one-hot encoding \cite{kotsiantis2006handling}. However, this approach increases dimensionality, introduces sparsity, and fails to capture semantic relationships between categories \cite{micci2001preprocessing,zheng2018feature}. These limitations can hinder model performance, particularly in datasets with high-cardinality categorical features.

\subsection{Advancements in Encoding Categorical Variables}

To address these limitations, alternative encoding techniques have been explored. \cite{guo2016entity} proposed using entity embeddings to represent categorical variables in a continuous vector space, allowing models to capture relationships between categories more effectively. Their findings demonstrated that models using entity embeddings outperformed those relying on one-hot encoding in various classification tasks.

However, the adoption of such techniques in employee attrition prediction remains limited. Most studies continue to rely on standard encoding methods, overlooking the potential benefits of contextual embeddings for improving model performance.

\subsection{Deep Learning and Self-Attention for Tabular Data}

Recent advancements in deep learning have introduced models capable of handling tabular data while preserving feature relationships. Transformer-based architectures such as \textit{TabTransformer} \cite{huang2020tabtransformer} and \textit{SAINT} \cite{somepalli2021saint} leverage self-attention mechanisms to generate contextual embeddings, addressing the limitations of traditional encoding methods.

\cite{huang2020tabtransformer} demonstrated that \textit{TabTransformer} outperforms conventional machine learning models by capturing interactions between categorical and numerical features. Similarly, \cite{somepalli2021saint} introduced \textit{SAINT} (Self-Attention and Intersample Attention Transformer), which incorporates both self-attention and intersample attention mechanisms to model complex feature relationships.

While these models have shown promise in tabular data classification, their application in employee attrition prediction remains unexplored. \textit{SAINT}-generated contextual embeddings have the potential to enhance feature representation and mitigate information loss in traditional encoding methods. Additionally, integrating these embeddings with tree-based models such as \textit{XGBoost} and \textit{LightGBM} may offer a hybrid approach that combines the strengths of both deep learning and conventional machine learning techniques.

\subsection{Research Gap}

Despite the advancements in deep learning for tabular data, the integration of transformer-based embeddings with tree-based classifiers remains an open research area, particularly in the context of employee attrition prediction. Existing studies primarily focus on handling class imbalance and evaluating model performance, with limited exploration of encoding strategies that preserve categorical feature relationships.

This study addresses this gap by investigating whether \textit{SAINT}-generated contextual embeddings can improve the predictive performance of tree-based models for employee attrition. By bridging the gap between deep learning and traditional classifiers, we aim to provide empirical insights into the effectiveness of transformer-based feature extraction in structured datasets.

\section{Method}

This section outlines the data used for our study, the exploratory data analysis~(EDA) conducted to understand its structure, the preprocessing steps applied to prepare it for modeling, and the experimental setup for evaluating different predictive models. We also describe the \textit{SAINT} model architecture and its integration with tree-based classifiers, as well as our approach to assessing interpretability through SHAP analysis.

\subsection{Dataset Description}
We use the publicly available Employee Attrition Dataset, which contains 74,498 employee records. The dataset is split into 59,598 samples for training and 14,900 for testing. Each record captures a mix of demographic, job-related, and organizational attributes that are potentially relevant for attrition prediction. Table~\ref{tab:features} provides a structured overview of the dataset features. The dataset is publicly accessible at the project repository:~\url{https://github.com/DerraziAdil/Importance_SAINT_ContextualEmbeddings}.

\begin{table}[h]
    \centering
    \caption{Summary of dataset features used for employee attrition prediction, categorized into demographic, job-related, and organizational attributes.}
    \label{tab:features}
    \resizebox{\textwidth}{!}{%
        \begin{tabular}{lll}
            \toprule
            \textbf{Feature Name} & \textbf{Type} & \textbf{Description} \\
            \midrule
            \multicolumn{3}{l}{\textbf{Demographic Attributes}} \\
            \midrule
            Age & Numerical & Employee's age (18 to 60 years) \\
            Gender & Binary & Male (1) or Female (0) \\
            Marital Status & Categorical & Divorced, Married, or Single \\
            Education Level & Categorical & Highest education level attained \\
            \midrule
            \multicolumn{3}{l}{\textbf{Job-Related Factors}} \\
            \midrule
            Job Role & Categorical & Employee's department (e.g., Finance, Tech) \\
            Monthly Income & Numerical & Monthly salary in dollars \\
            Years at Company & Numerical & Duration of employment \\
            Performance Rating & Categorical & Performance evaluation score \\
            Work-Life Balance & Categorical & Employee’s perceived balance \\
            Job Satisfaction & Categorical & Employee’s job satisfaction level \\
            Remote Work & Binary & Yes (1) or No (0) \\
            \midrule
            \multicolumn{3}{l}{\textbf{Organizational Factors}} \\
            \midrule
            Company Size & Categorical & Small, Medium, or Large \\
            Company Reputation & Categorical & Employee’s perception of reputation \\
            Employee Recognition & Categorical & Recognition level received \\
            \midrule
            \textbf{Attrition} & \textbf{Binary (Target)} & Left (1) or Stayed (0) \\
            \bottomrule
        \end{tabular}%
    }
\end{table}

\subsection{Exploratory Data Analysis}
We performed EDA to identify patterns and potential issues within the dataset. The dataset contains no missing values, so no imputation was required. 

Descriptive statistics showed that numerical features such as \texttt{Age}, \texttt{Years at Company}, and \texttt{Monthly Income} had reasonable distributions without extreme outliers. Categorical attributes, such as job role and work-life balance, were examined to understand workforce composition and potential attrition drivers. The dataset was also found to be relatively balanced, with 52.5\% of employees staying and 47.5\% leaving, ensuring that class imbalance handling was unnecessary.

To identify relationships between features, we computed Pearson correlation coefficients. As expected, \texttt{Years at Company} correlated positively with \texttt{Age} and \texttt{Company Tenure}, while other features showed low correlation, indicating their relative independence. A full correlation heatmap and outlier detection analysis are available in Appendix~\ref{app:eda_figures}.

\subsection{Data Preprocessing}
Our preprocessing pipeline was designed to optimize model performance while preventing data leakage. Since no missing values were detected, all records were retained. 

Binary features (e.g., \texttt{Gender}, \texttt{Remote Work}) were encoded as 0 or 1. Categorical features were handled differently based on the modeling approach:

\begin{itemize}
    \item \textbf{Tree-based models (XGBoost, LightGBM):} One-hot encoding was applied using \texttt{OneHotEncoder} from \texttt{scikit-learn}.
    \item \textbf{SAINT model:} Ordinal encoding was used to map each category to a unique integer, aligning with \textit{SAINT}'s embedding mechanism.
    \item \textbf{Hybrid models (SAINT-XGBoost, SAINT-LightGBM):} Ordinal encoding was first applied, and \textit{SAINT}-generated embeddings were used as additional inputs for tree-based classifiers.
\end{itemize}

Numerical features were standardized using z-score normalization with \texttt{StandardScaler}, applied separately within each cross-validation fold to avoid data leakage.

\subsection{Experimental Design}
\begin{figure}[h]
    \centering
    \includegraphics[width=\textwidth]{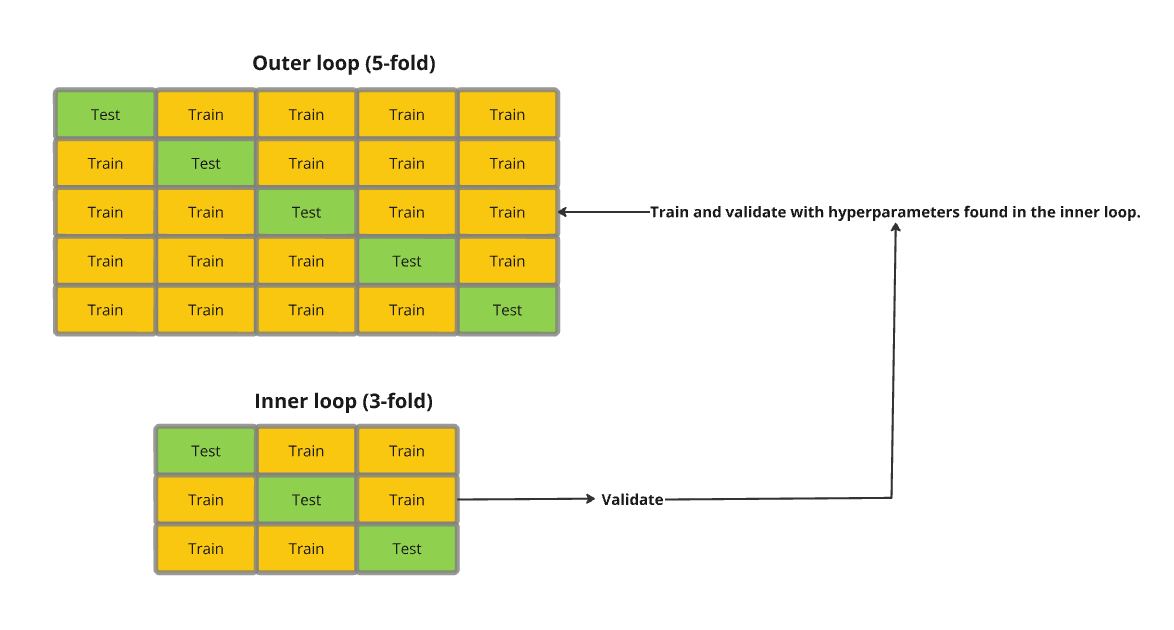}
    \caption{Nested cross-validation setup. The outer loop evaluates model performance, while the inner loop fine-tunes hyperparameters.}
    \label{fig:nested}
\end{figure}

We employed a nested cross-validation framework to optimize and evaluate model performance. The outer loop used five-fold cross-validation, while the inner loop performed three-fold cross-validation for hyperparameter tuning. This approach ensured robust performance estimation while mitigating overfitting (Figure~\ref{fig:nested}).

\subsection{SAINT Model Architecture}
\begin{figure}[h]
    \centering
    \includegraphics[scale=0.5]{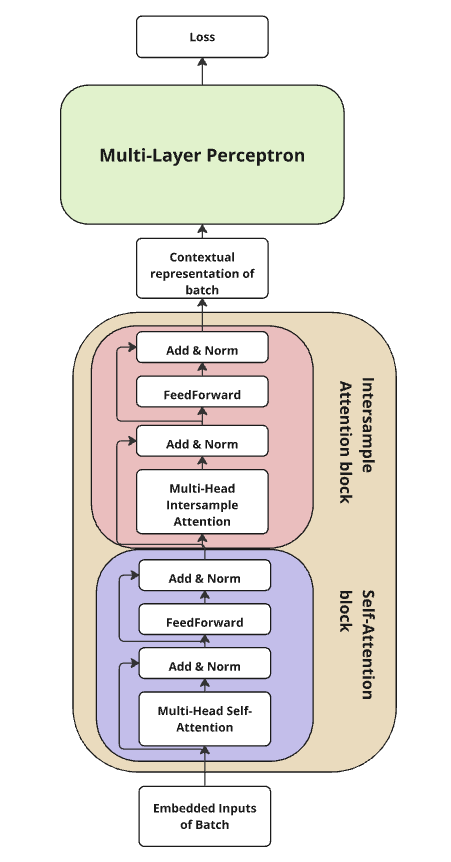}
    \caption{\textit{SAINT} model architecture. Features are transformed into embeddings, processed through self-attention and intersample attention blocks, and passed through a feedforward network for classification.}
    \label{fig:saint_architecture}
\end{figure}

\textit{SAINT} is a transformer-based model designed for tabular data~\cite{somepalli2021saint}. It integrates self-attention and intersample attention mechanisms to enhance feature representation. The model's architecture consists of embedding layers, attention blocks, and a feedforward network (Figure~\ref{fig:saint_architecture}).

\subsection{Evaluation and Interpretability}
Model performance was primarily assessed using the ROC-AUC metric, along with precision, recall, and F1-score. To determine the statistical significance of performance differences, DeLong’s test was used to compare ROC curves~\cite{robin2011proc}.

To enhance interpretability, we employed SHAP (SHapley Additive exPlanations) analysis to quantify feature contributions. This analysis compared the importance of features across different models, helping to assess whether hybrid pipelines improved interpretability.

\subsection{Implementation Details}
Hyperparameters were optimized using grid search within the inner loop of nested cross-validation. The best configurations, determined based on ROC-AUC scores, were then applied to the final models. Detailed hyperparameter settings are available in Appendix~\ref{app:hyperparameters}.

\section{Results}

This section presents the findings of our study, focusing on the impact of \emph{SAINT}-generated embeddings on the predictive performance of tree-based models for employee attrition prediction. We compare hybrid pipelines integrating \emph{SAINT} with tree-based models (\emph{XGBoost} and \emph{LightGBM}) against their standalone counterparts. The primary evaluation metric is ROC-AUC, complemented by precision, recall, and F1-score for a more comprehensive assessment. Also, we evaluate computational efficiency through training times and conduct error analysis to examine model generalization and misclassification patterns.

\subsection{Performance Metrics}
Table~\ref{tab:test_metrics} summarizes the test set performance for all models, including ROC-AUC, precision, recall, F1-score, and execution time. Additional plots illustrating ROC-AUC scores and performance metrics across training, validation, and test sets can be found in Appendix~\ref{app:e}.

\begin{table}[htbp]
\centering
\caption{Test Set Performance Metrics for All Models}
\label{tab:test_metrics}
\resizebox{\textwidth}{!}{%
\begin{tabular}{lccccc}
\toprule
\textbf{Model} & \textbf{ROC-AUC} & \textbf{Precision} & \textbf{Recall} & \textbf{F1-Score} & \textbf{Time (s)} \\
\midrule
\emph{SAINT-XGBoost}    & 0.8233 & 0.7175 & 0.7403 & 0.7287 & 91.12 \\
\emph{SAINT-LightGBM}   & 0.8431 & 0.6941 & 0.6871 & 0.6906 & 92.53 \\
\emph{SAINT}        & 0.8431 & 0.6748 & \textbf{0.8468} & \textbf{0.7511} & 89.32 \\
\emph{XGBoost}      & \textbf{0.8529} & \textbf{0.7461} & 0.7446 & 0.7453 & 23.40 \\
\emph{LightGBM}     & 0.8521 & 0.7454 & 0.7472 & 0.7463 & \textbf{18.34} \\
\bottomrule
\end{tabular}%
}
\end{table}

\subsection{Model Performance Analysis}

\subsubsection{ROC-AUC Comparison}
ROC-AUC evaluates a model’s ability to distinguish between employees who stay and those who leave. The \emph{XGBoost} model attained the highest ROC-AUC score (0.8529), followed closely by \emph{LightGBM} (0.8521). The \emph{SAINT} model achieved a slightly lower ROC-AUC of 0.8431, comparable to \emph{SAINT-LightGBM}, but both hybrid models underperformed compared to their standalone tree-based counterparts. 

DeLong’s test showed no significant difference between hybrid models and SAINT (p = 0.1495, p = 0.1977). However, tree-based models significantly outperformed hybrids (p $<$ 0.05). Detailed results of these statistical comparisons are provided in Table~\ref{tab:stat_tests}.

\begin{table}[htbp]
\centering
\caption{Statistical Comparison of ROC-AUC Scores (DeLong)}
\label{tab:stat_tests}
\resizebox{\textwidth}{!}{%
\begin{tabular}{lcccccl}
\toprule
\textbf{Model Pair} & \textbf{AUC Model 1} & \textbf{AUC Model 2} & \textbf{p-value} & \textbf{Log10(p-value)} & \textbf{Significance} \\
\midrule
\emph{SAINT-XGBoost} vs. \emph{SAINT-LightGBM} & 0.8448 & 0.8431 & 1.4948e-01 & -0.8254 & Not Significant \\
\emph{SAINT-XGBoost} vs. \emph{LightGBM}       & 0.8448 & 0.8521 & 3.7356e-10 & -9.4276 & Significant \\
\emph{SAINT-XGBoost} vs. \emph{SAINT}          & 0.8448 & 0.8431 & 1.9773e-01 & -0.7039 & Not Significant \\
\emph{SAINT-XGBoost} vs. \emph{XGBoost}        & 0.8448 & 0.8529 & 1.4537e-12 & -11.8375 & Significant \\
\emph{SAINT-LightGBM} vs. \emph{LightGBM}      & 0.8431 & 0.8521 & 6.8891e-13 & -12.1618 & Significant \\
\emph{SAINT-LightGBM} vs. \emph{SAINT}         & 0.8431 & 0.8431 & 9.9714e-01 & -0.0012 & Not Significant \\
\emph{SAINT-LightGBM} vs. \emph{XGBoost}       & 0.8431 & 0.8529 & 4.2089e-15 & -14.3758 & Significant \\
\emph{LightGBM} vs. \emph{SAINT}               & 0.8521 & 0.8431 & 2.5447e-12 & -11.5944 & Significant \\
\emph{LightGBM} vs. \emph{XGBoost}             & 0.8521 & 0.8529 & 3.7239e-03 & -2.4290 & Significant \\
\emph{SAINT} vs. \emph{XGBoost}                & 0.8431 & 0.8529 & 6.8020e-15 & -14.1674 & Significant \\
\bottomrule
\end{tabular}%
}
\end{table}

\subsubsection{Precision, Recall, and F1-Score}
While \emph{SAINT} achieved the highest recall (0.8468), it had the lowest precision (0.6748), suggesting a higher rate of false positives. \emph{XGBoost} demonstrated balanced performance with the highest precision (0.7461) and a competitive F1-score (0.7453), while \emph{LightGBM} showed a similar pattern. Hybrid models exhibited lower scores across all three metrics, indicating that the additional embeddings did not enhance tree-based models’ predictive capabilities.

\subsubsection{Computational Efficiency}
Training and inference times were also considered. \emph{LightGBM} was the most efficient model, completing execution in 18.34 seconds, while \emph{XGBoost} took 23.40 seconds. In contrast, the \emph{SAINT} model required significantly more time (89.32 seconds), with hybrid pipelines taking even longer (91.12 and 92.53 seconds for \emph{SAINT-XGBoost} and \emph{SAINT-LightGBM}, respectively). The additional computational burden of \emph{SAINT}-generated embeddings did not result in proportional performance gains, making standalone tree-based models more efficient choices.

\subsection{Error Analysis}

\subsubsection{Confusion Matrices}
Confusion matrices were analyzed to further understand model errors. The \emph{XGBoost} model correctly classified 11,322 instances, with 6,086 true negatives and 5,236 true positives, whereas \emph{SAINT-XGBoost} exhibited higher false positives (2,050) and false negatives (1,826), indicating reduced accuracy. 

Similarly, \emph{SAINT} correctly identified more true positives (5,955) but had a higher false positive rate (2,870), leading to lower precision. Standalone tree-based models demonstrated more balanced classifications, with fewer misclassifications overall.

\subsubsection{Generalization Performance}
Cross-validation results in Table~\ref{tab:cv_metrics} show that \emph{XGBoost} and \emph{LightGBM} achieved the highest mean ROC-AUC scores (0.8506 and 0.8499), with low variance across folds, suggesting strong generalization. In contrast, hybrid models displayed greater performance variation, further reinforcing that \emph{SAINT}-generated embeddings did not enhance model robustness.

\begin{table}[htbp]
\centering
\caption{Cross-Validation ROC-AUC Scores}
\label{tab:cv_metrics}
\resizebox{\textwidth}{!}{%
\begin{tabular}{lcccc}
\toprule
\textbf{Model} & \textbf{ROC-AUC (Test Set)} & \textbf{Delta (Test - Validation)} & \textbf{Mean ROC-AUC (Validation Set)} & \textbf{Standard Deviation (Validation Set)} \\
\midrule
\emph{SAINT-XGBoost } & 0.8233 & \textbf{-0.0151} & 0.8384 & 0.0031 \\
\emph{SAINT-LightGBM } & 0.8431 & -0.0016 & 0.8347 & 0.0034 \\
\emph{SAINT }       & 0.8431 & +0.0046 & 0.8385 & 0.0034 \\
\emph{XGBoost }     & \textbf{0.8529} & +0.0023 & \textbf{0.8506} & \textbf{0.0043} \\
\emph{LightGBM }    & 0.8521 & +0.0022 & 0.8499 & 0.0042 \\
\bottomrule
\end{tabular}%
}
\end{table}

\subsection{SHAP Analysis for Model Interpretability}
\begin{figure}[!ht]
    \centering

    % === Absolute SHAP Values ===
    \begin{subfigure}{0.48\linewidth}
        \includegraphics[width=\linewidth]{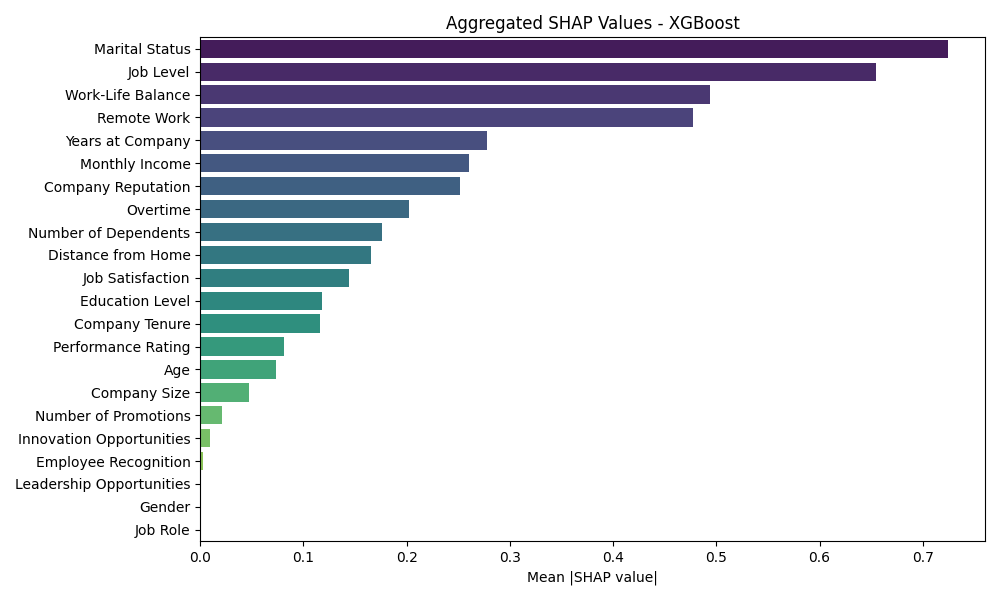}
        \caption{XGBoost}
        \label{fig:xgboost_shap_abs}
    \end{subfigure}
    \hfill
    \begin{subfigure}{0.48\linewidth}
        \includegraphics[width=\linewidth]{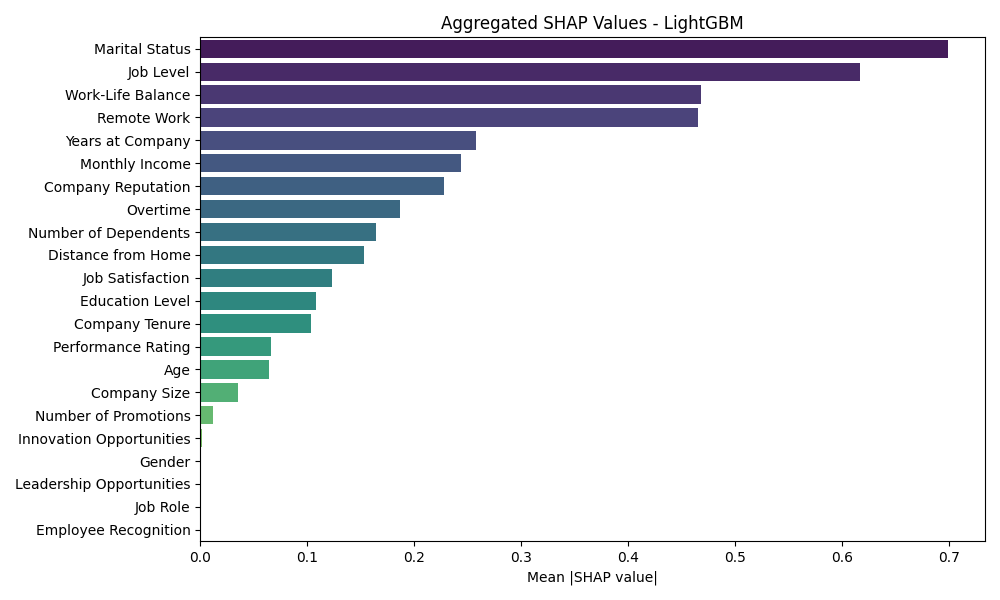}
        \caption{LightGBM}
        \label{fig:lgbm_shap_abs}
    \end{subfigure}

    \vspace{1em}

    \begin{subfigure}{0.48\linewidth}
        \includegraphics[width=\linewidth]{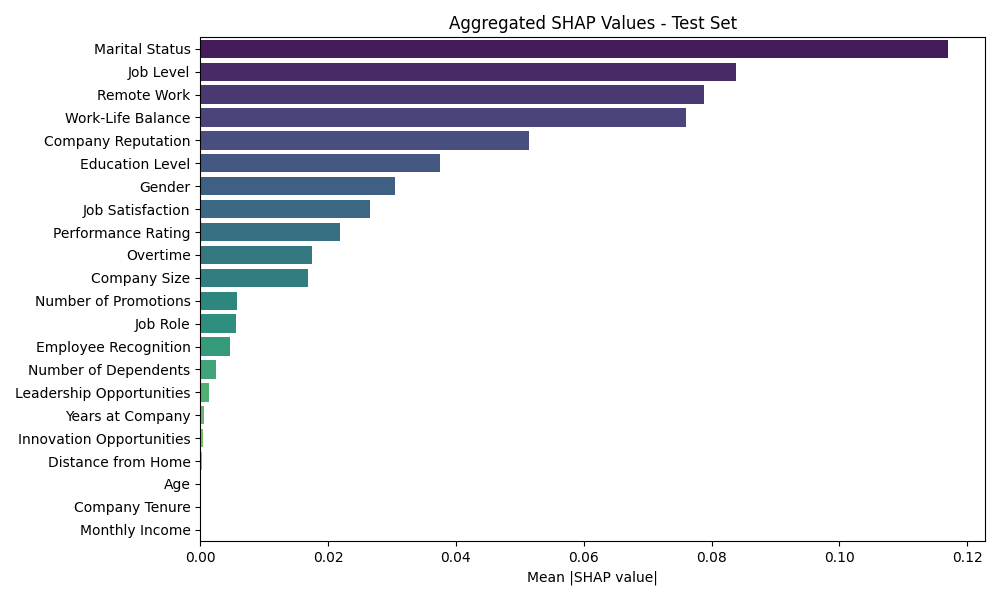}
        \caption{SAINT}
        \label{fig:saint_shap_abs}
    \end{subfigure}
    \hfill
    \begin{subfigure}{0.48\linewidth}
        \includegraphics[width=\linewidth]{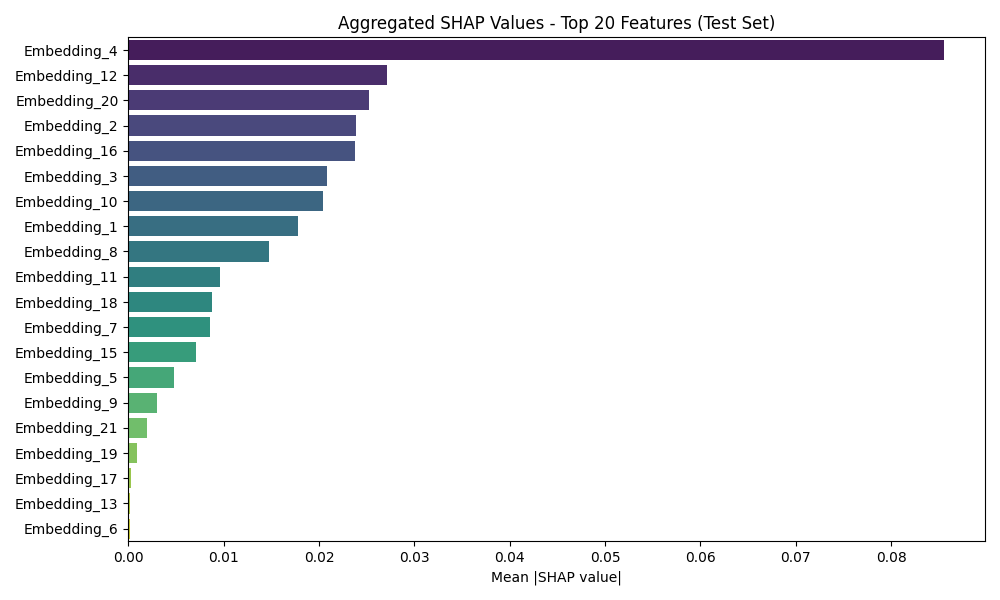}
        \caption{XGBoost-SAINT}
        \label{fig:SHAP_SAINT-XGBOOST_ABS}
    \end{subfigure}

    \vspace{1em}

    \begin{subfigure}{0.6\linewidth}
        \includegraphics[width=\linewidth]{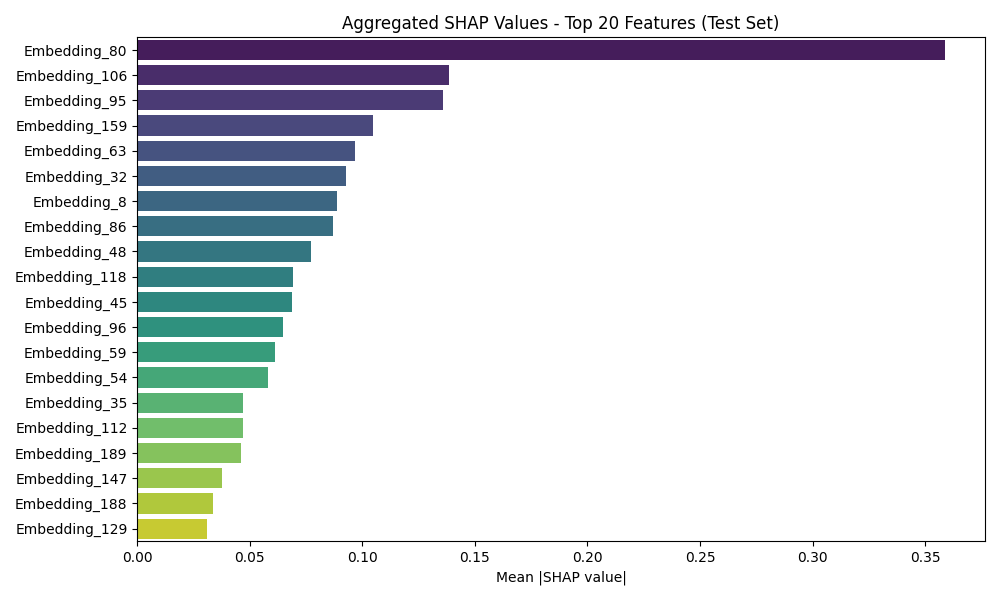}
        \caption{LightGBM-SAINT}
        \label{fig:SHAP_SAINT-LGBM_ABS}
    \end{subfigure}

    \caption{Absolute SHAP values for various models.}
    \label{fig:shap_abs_all}
\end{figure}

SHAP analysis was performed to evaluate feature importance. Figures~\ref{fig:xgboost_shap_abs}, \ref{fig:lgbm_shap_abs}, and \ref{fig:saint_shap_abs} illustrate the most influential features for each model. 

Tree-based models identified \texttt{Marital Status}, \texttt{Job Level}, and \texttt{Work-Life Balance} as the most impactful features, whereas \emph{SAINT} assigned relatively higher importance to categorical and binary features. The hybrid models, however, posed interpretability challenges as their SHAP values pertain to abstracted embeddings rather than original features, limiting direct insights.

Overall, the SHAP analysis suggests that while \emph{SAINT} captures meaningful feature relationships, tree-based models provide clearer and more interpretable importance scores, reinforcing their practical utility for decision-making in employee attrition prediction.

\section{Discussion}
This study examined the impact of \emph{SAINT}-generated embeddings on the predictive performance of tree-based models for employee attrition prediction. The central research question explored whether integrating \emph{SAINT} embeddings enhances predictive accuracy compared to traditional encoding methods in tree-based models. To address this, three key aspects were investigated: predictive performance, generalizability, and interpretability.

\subsection{Predictive Performance}
The first objective was to compare predictive performance across different model configurations. Specifically, we analyzed how the hybrid pipelines (\emph{SAINT-XGBoost} and \emph{SAINT-LightGBM}) performed relative to standalone models (\emph{XGBoost}, \emph{LightGBM}, and \emph{SAINT}) using ROC-AUC, precision, recall, and F1-score as evaluation metrics.

The results showed that the standalone tree-based models, \emph{XGBoost} and \emph{LightGBM}, achieved the highest ROC-AUC scores (0.8529 and 0.8521, respectively), outperforming the standalone \emph{SAINT} model (0.8431). The hybrid pipelines did not improve upon these results, with ROC-AUC scores of 0.8233 and 0.8431 for \emph{SAINT-XGBoost} and \emph{SAINT-LightGBM}, respectively. Statistical analysis using DeLong’s test confirmed that these differences were significant, indicating that integrating \emph{SAINT} embeddings did not enhance predictive performance.

Beyond ROC-AUC, \emph{SAINT} exhibited the highest recall (0.8468) and F1-score (0.7511), suggesting strong sensitivity to positive cases. However, this came at the cost of lower precision (0.6748). In contrast, \emph{XGBoost} achieved the highest precision (0.7461), striking a more balanced performance. The hybrid pipelines showed weaker performance across all these metrics, reinforcing the conclusion that \emph{SAINT}-generated embeddings did not provide added value when combined with tree-based models.

These findings align with previous research highlighting that tree-based models struggle with dense, high-dimensional embeddings. Unlike neural networks, which learn feature hierarchies, tree-based models rely on axis-aligned splits, which are not optimized for processing continuous vector representations. As a result, they may fail to leverage the rich semantic information captured in SAINT-generated embeddings. Prior studies \cite{shwartz2022tabular,nguyen2019tree} have demonstrated that tree-based models excel with structured tabular data but face challenges when handling dense embeddings. Given that \emph{SAINT} generates contextualized representations, its embeddings may not be well-suited for tree-based architectures, which operate best with sparse tabular inputs.

A notable observation is that while the hybrid pipelines underperformed relative to the standalone tree-based models, they performed comparably to standalone \emph{SAINT}. This is somewhat unexpected, as prior literature suggests that tree-based models do not effectively leverage high-dimensional embeddings \cite{nguyen2019tree,Lin2023TreeIndex}. One explanation is that \emph{SAINT} embeddings retained sufficient predictive information but were not optimized for tree-based classifiers. Another possibility is that the variance captured in the embeddings limited their effectiveness, as the tree-based classifiers struggled to fully exploit the learned representations.

\subsection{Generalizability}
To assess model generalizability, we compared cross-validation ROC-AUC scores with test set performance. The standalone tree-based models exhibited minimal differences between validation and test performance (\emph{XGBoost}: +0.0023, \emph{LightGBM}: +0.0022), indicating strong generalizability. Similarly, the \emph{SAINT} model maintained good generalizability with a delta of +0.0046.

In contrast, the \emph{SAINT-XGBoost} hybrid pipeline showed a notable performance drop (-0.0151), suggesting overfitting during training. The \emph{SAINT-LightGBM} pipeline exhibited a smaller decrease (-0.0016), indicating relatively stable generalization. These results further emphasize that integrating \emph{SAINT} embeddings with tree-based models does not improve––and may even detract from––generalizability compared to standalone models.

\subsection{Interpretability}
Interpretability is a critical consideration in HR analytics, where model transparency is essential. Using SHAP analysis, we examined feature importance in standalone and hybrid models.

For standalone models, SHAP values provided clear insights. The tree-based models and \emph{SAINT} identified similar top features influencing attrition predictions. However, \emph{SAINT} placed greater emphasis on categorical and binary features, likely due to its attention mechanisms capturing complex feature interactions.

Hybrid models posed interpretability challenges since SHAP values were computed on embeddings rather than original features. While SAINT captures complex relationships, its integration with tree-based classifiers obscures feature-level insights. For HR applications, where decision-making transparency is crucial, this reduced interpretability may hinder adoption.

\subsection{Practical and Societal Implications}
Accurate attrition prediction can help organizations implement proactive retention strategies. However, the findings suggest that simpler, interpretable tree-based models outperform complex hybrid architectures while maintaining transparency. This reinforces the importance of balancing predictive performance with interpretability in HR analytics.

Ethical considerations are also paramount. The use of opaque models increases the risk of unintended biases. Transparent models facilitate bias detection and mitigation, ensuring fairness in employee retention strategies. The study underscores the need for explainable AI in HR contexts to foster trust and responsible AI adoption.

\subsection{Limitations and Future Directions}
Our \textit{SAINT} hyperparameter tuning was constrained by computational resources. Future studies could adopt more extensive optimization techniques such as Bayesian optimization or reinforcement-based search to determine optimal settings (for example, attention heads, embedding dimension, and dropout rate). This could uncover configurations that yield stronger predictive performance. Similarly, more rigorous tuning of tree-based classifiers, including larger ensembles and advanced regularization, might clarify whether and how dense contextual embeddings can complement the way these models split features.

A central limitation is the difficulty of interpreting the hybrid pipelines. Once numeric embeddings replace the original categorical features, direct feature-level insights become less clear. Future work should focus on methods that map these learned embeddings back to their original features, making the representations more transparent. Visual analytics, such as \textit{t-SNE} or \textit{UMAP} plots of \textit{SAINT} embeddings, could help illustrate how different feature values group together in the embedding space, thereby helping practitioners see contextual relationships and refine HR policies.

Our hybrid approach used \textit{SAINT} strictly as a feature extractor before passing data to \textit{XGBoost} or \textit{LightGBM}. Future studies could investigate deeper fusion strategies in which \textit{SAINT} and tree-based models share intermediate layers or loss functions, potentially capturing a richer interplay between attention-driven embeddings and axis-aligned splits. Joint training objectives, where gradient updates propagate back into \textit{SAINT}’s embedding layers, may help align these learned representations with the decision paths in tree-based ensembles.

This study was conducted using a single HR dataset. Given the variety of real-world HR systems with differences in region, industry, company size, and workforce structure, validating these methods on diverse datasets is essential. Examining data with different distributions of categorical and numerical attributes would help determine whether certain conditions favor transformer-based embeddings, thereby guiding organizations in deciding when to invest in deep learning solutions for employee attrition.

\subsection{Conclusion}
This study examined the impact of \emph{SAINT}-generated embeddings on tree-based models for employee attrition prediction. The findings indicate that hybrid pipelines did not enhance predictive performance. Instead, standalone tree-based models (\emph{XGBoost} and \emph{LightGBM}) outperformed both the hybrid pipelines and the standalone \emph{SAINT} model in terms of ROC-AUC, precision-recall balance, and generalizability.

These results suggest that tree-based models remain the preferred approach for structured tabular data in attrition prediction. They offer not only superior predictive performance but also greater interpretability, which is crucial for HR decision-making. The challenges in hybrid model interpretability highlight the need for further research on enhancing compatibility between deep learning embeddings and traditional classifiers. Future studies should explore embedding interpretability, optimize model architectures, and validate findings across diverse datasets to assess their robustness in real-world applications.

\section*{Appendix A – EDA Figures}  
\label{app:eda_figures}
This section presents a few example visualizations from the EDA. While only selected figures are included here, a full set of visualizations, including box plots for all numerical features, is available in the project repository at \url{https://github.com/DerraziAdil/Importance_SAINT_ContextualEmbeddings}.

\begin{figure}[H]
    \centering

    \begin{subfigure}{0.32\textwidth}
        \includegraphics[width=\linewidth]{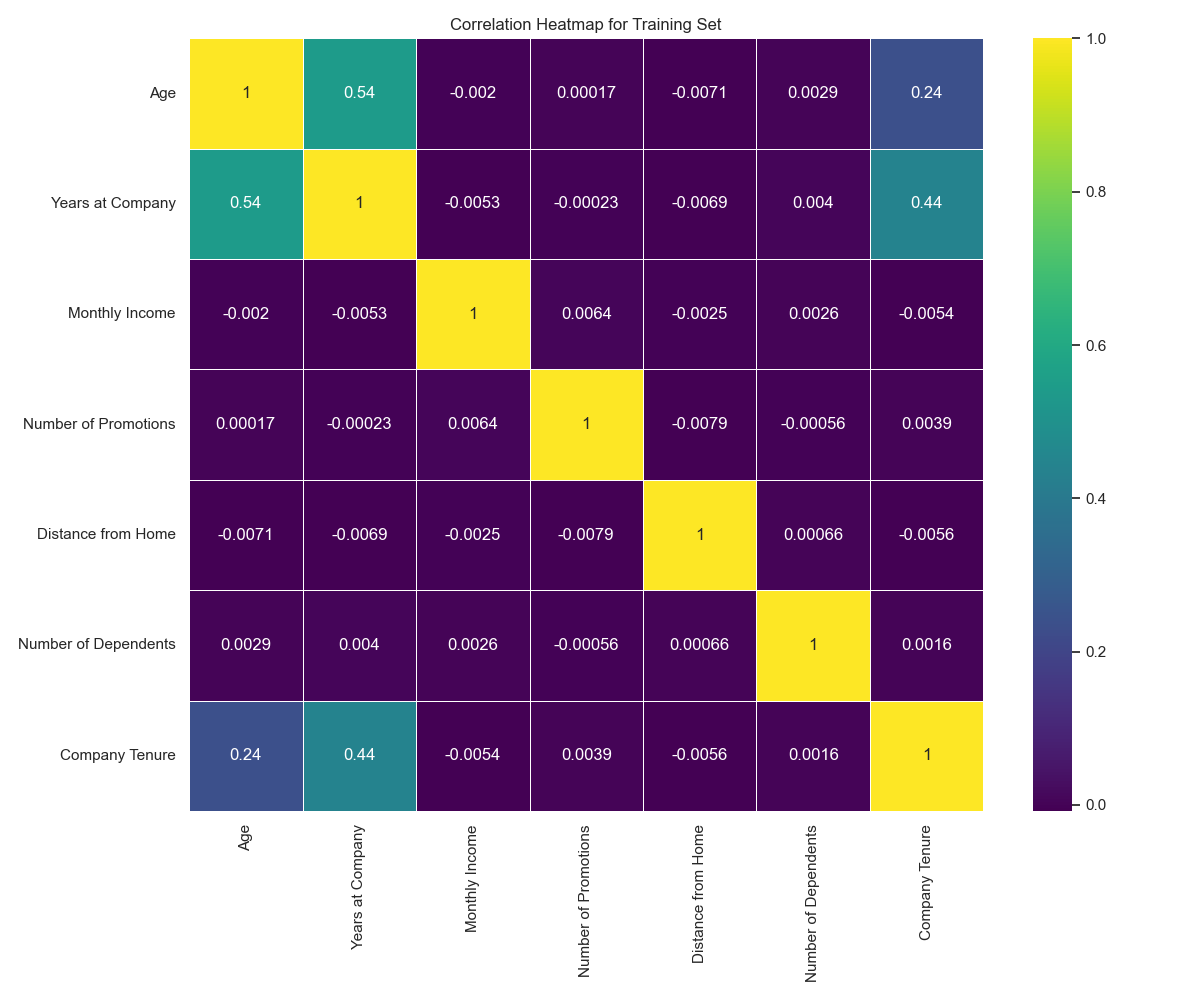}
        \caption{Correlation heatmap (Train)}
        \label{fig:correlation-heatmap-train}
    \end{subfigure}
    \hfill
    \begin{subfigure}{0.32\textwidth}
        \includegraphics[width=\linewidth]{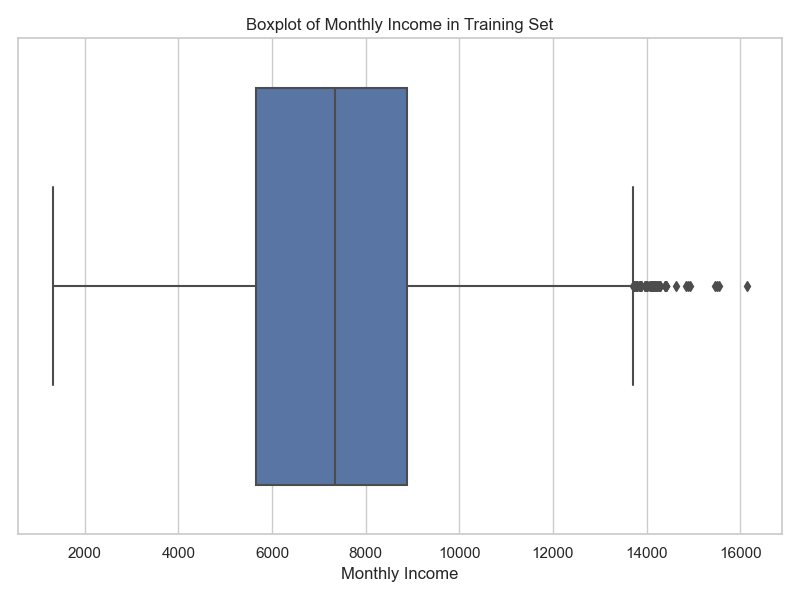}
        \caption{Monthly Income (Train)}
        \label{fig:box-income-train}
    \end{subfigure}
    \hfill
    \begin{subfigure}{0.32\textwidth}
        \includegraphics[width=\linewidth]{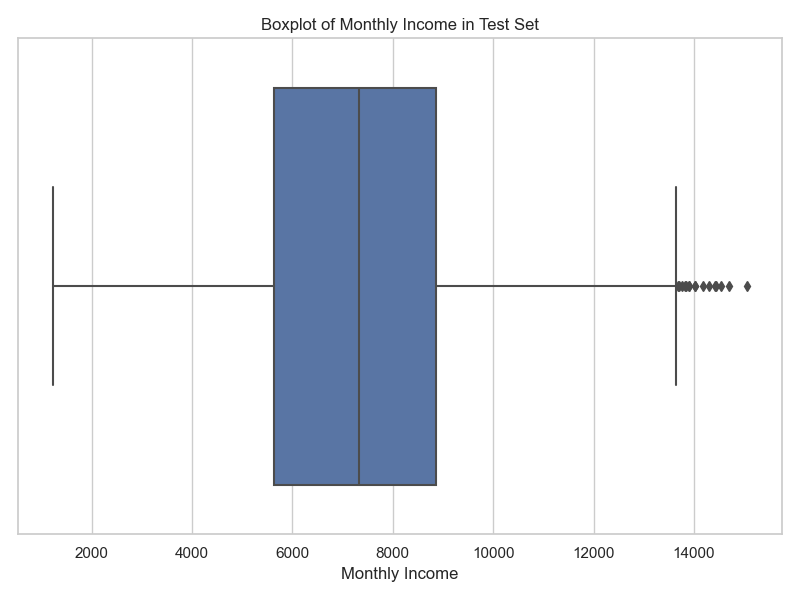}
        \caption{Monthly Income (Test)}
        \label{fig:box-income-test}
    \end{subfigure}

    \caption{EDA visualizations for correlation and income distributions.}
    \label{fig:eda_visuals}
\end{figure}

\FloatBarrier

\section*{Appendix B – Hyperparameter Tuning Grids}
\label{app:hyperparameters}

This appendix provides the hyperparameter tuning grids explored for each model.

\begin{table}[H]
    \centering
    \caption{Hyperparameter tuning grids for \textit{XGBoost} and \textit{LightGBM}.}
    \begin{minipage}{0.48\textwidth}
        \centering
        \textbf{XGBoost}
        \vspace{0.5em}
        \adjustbox{max width=\linewidth}{%
        \begin{tabular}{|l|l|}
            \hline
            \textbf{Parameter} & \textbf{Values} \\ \hline
            Estimators & [300, 400, 500] \\ \hline
            Max Depth & [2, 3, 4] \\ \hline
            LR & [0.05, 0.01] \\ \hline
            Subsample & 0.9 \\ \hline
            Colsample & [0.8, 1.0] \\ \hline
            Min Child Wt & [3, 4] \\ \hline
            Reg Lambda & [1, 2] \\ \hline
            Reg Gamma & 0 \\ \hline
        \end{tabular}
        }
    \end{minipage}
    \hfill
    \begin{minipage}{0.48\textwidth}
        \centering
        \textbf{LightGBM}
        \vspace{0.5em}
        \adjustbox{max width=\linewidth}{%
        \begin{tabular}{|l|l|}
            \hline
            \textbf{Parameter} & \textbf{Values} \\ \hline
            LR & [0.1, 0.05] \\ \hline
            Leaves & [5, 10, 20] \\ \hline
            Estimators & [200, 300, 400] \\ \hline
            Subsample & 0.8 \\ \hline
            Colsample & 0.8 \\ \hline
            Reg Alpha & 0.1 \\ \hline
            Reg Lambda & 5 \\ \hline
        \end{tabular}
        }
    \end{minipage}
\end{table}

\begin{table}[H]
    \centering
    \caption{Hyperparameter tuning grid for \textit{SAINT}.}
    \adjustbox{max width=\textwidth}{%
    \begin{tabular}{|l|l|}
        \hline
        \textbf{Parameter} & \textbf{Values Tested} \\ \hline
        Embedding Dimension & [16, 32, 64] \\ \hline
        Hidden Dimension & [32, 64, 128] \\ \hline
        Attention Heads & [2, 4] \\ \hline
        Transformer Layers & [2, 4] \\ \hline
        Dropout Rate & [0.1, 0.5] \\ \hline
        Learning Rate & 0.001 \\ \hline
        Batch Size & [50, 100] \\ \hline
    \end{tabular}
    }
    \label{tab:saint_params}
\end{table}

\FloatBarrier

\section*{Appendix C – Best Model Configurations}
\label{app:best_parameters}

The following table presents the best hyperparameter configuration for the standalone \textit{SAINT} model.

\begin{table}[H]
    \centering
    \caption{Best-performing hyperparameters for \textit{SAINT}.}
    \adjustbox{max width=\textwidth}{%
    \begin{tabular}{|l|l|}
        \hline
        \textbf{Parameter} & \textbf{Value} \\ \hline
        Embedding Dimension & 16 \\ \hline
        Hidden Dimension & 64 \\ \hline
        Attention Heads & 2 \\ \hline
        Layers & 2 \\ \hline
        Dropout Rate & 0.1 \\ \hline
        Learning Rate & 0.001 \\ \hline
        Batch Size & 100 \\ \hline
        Epochs & 1 \\ \hline
    \end{tabular}
    }
    \label{tab:saint_best_params}
\end{table}

\FloatBarrier

\section*{Appendix E – Additional Figures and Tables}
\label{app:e}

This appendix presents supplementary materials supporting the findings in the results section. It includes additional figures for ROC-AUC curves, performance metric comparisons across training, validation, and test sets, as well as confusion matrices and SHAP analysis details.

\subsection{ROC-AUC Curves}
The following figures illustrate the ROC-AUC curves for all models across training, validation, and test sets.

\begin{figure}[!ht]
    \centering

    \begin{subfigure}{0.48\linewidth}
        \includegraphics[width=\linewidth]{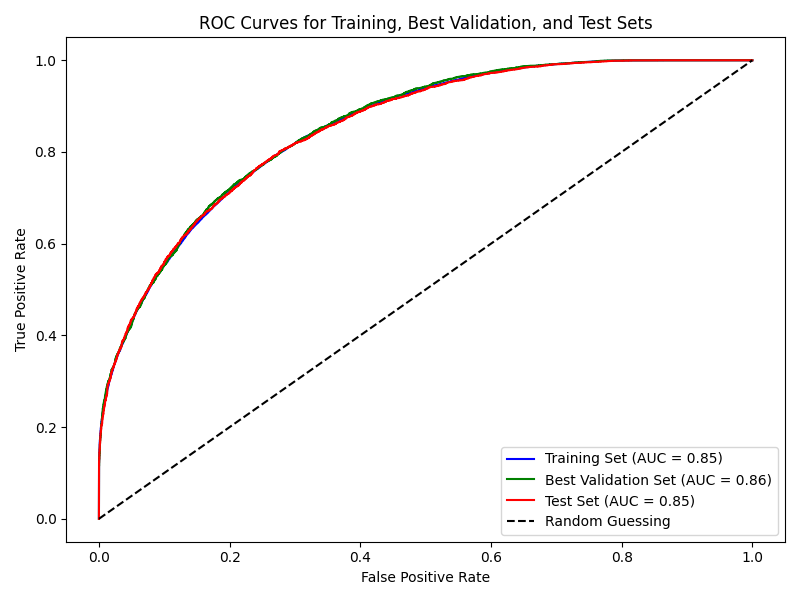}
        \caption{XGBoost}
        \label{fig:ROC_XGBoost}
    \end{subfigure}
    \hfill
    \begin{subfigure}{0.48\linewidth}
        \includegraphics[width=\linewidth]{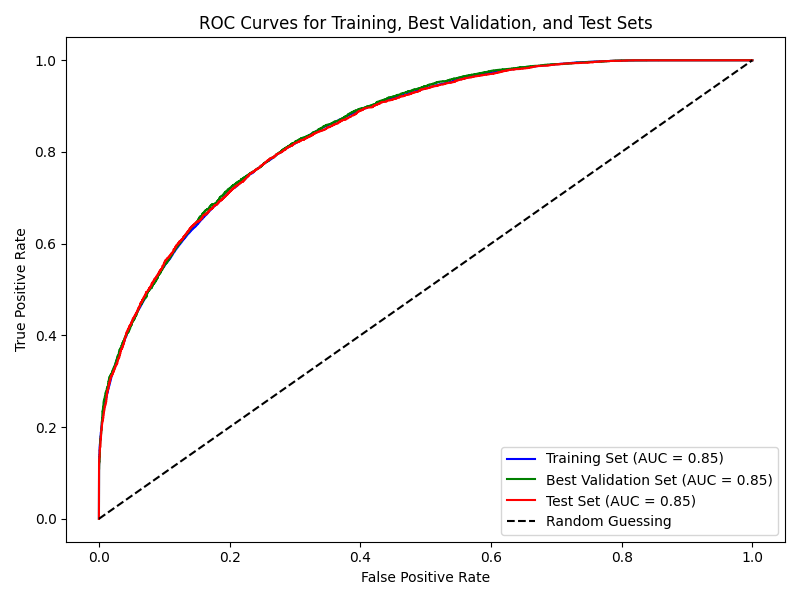}
        \caption{LightGBM}
        \label{fig:ROC_LightGBM}
    \end{subfigure}

    \vspace{1em}

    \begin{subfigure}{0.48\linewidth}
        \includegraphics[width=\linewidth]{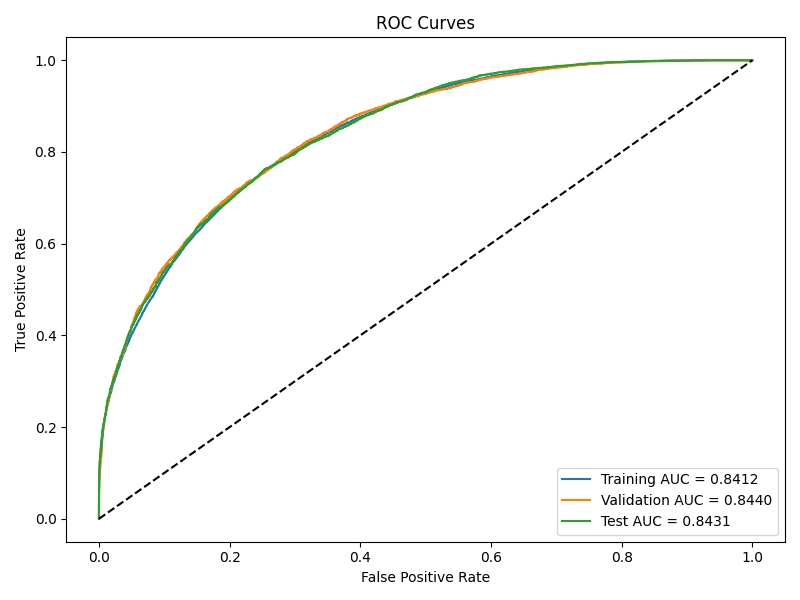}
        \caption{SAINT}
        \label{fig:ROC_SAINT}
    \end{subfigure}
    \hfill
    \begin{subfigure}{0.48\linewidth}
        \includegraphics[width=\linewidth]{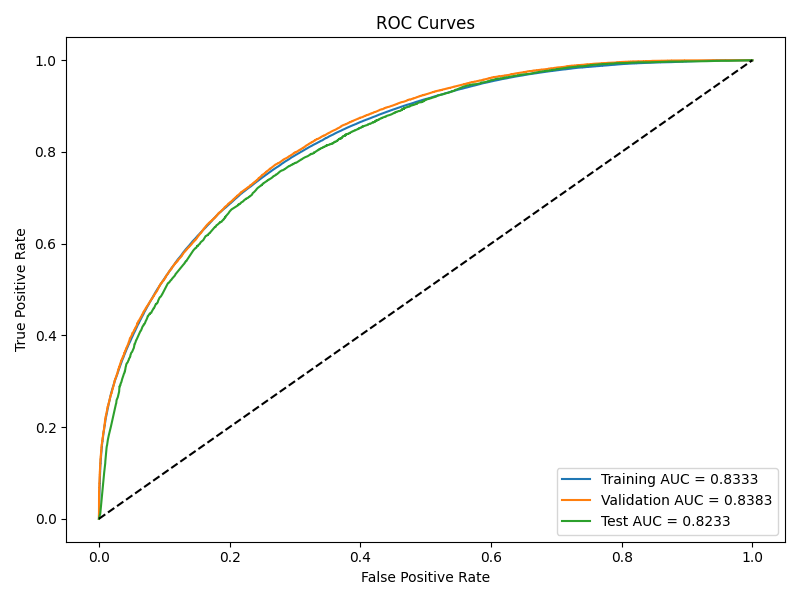}
        \caption{XGBoost-SAINT}
        \label{fig:ROC_XGBoost_SAINT}
    \end{subfigure}

    \vspace{1em}

    \begin{subfigure}{0.48\linewidth}
        \includegraphics[width=\linewidth]{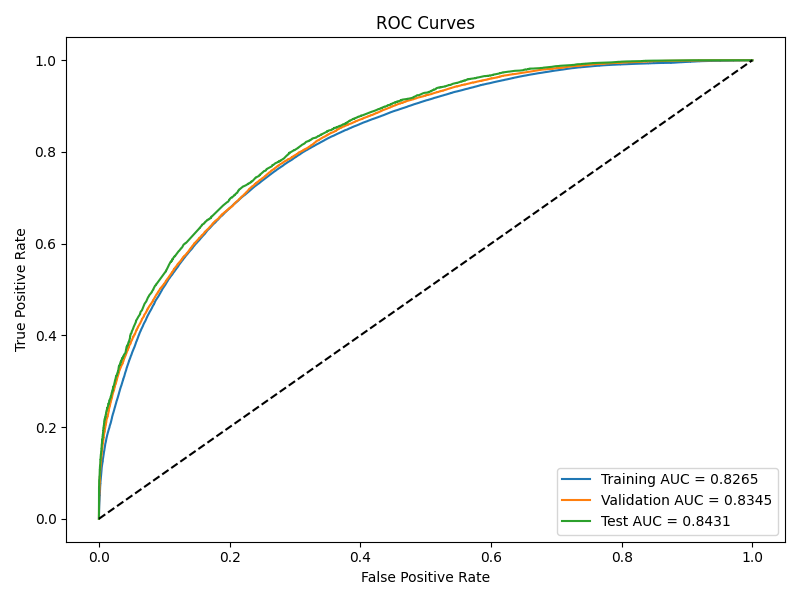}
        \caption{SAINT-LightGBM}
        \label{fig:ROC_SAINT_LightGBM}
    \end{subfigure}

    \caption{ROC Curves for Train, Validation, and Test Sets across different models}
    \label{fig:roc_all_models}
\end{figure}

\FloatBarrier

\subsection{Confusion Matrices}
To better understand misclassification patterns, confusion matrices for key models are provided below.

\begin{figure}[!ht]
    \centering

    \begin{subfigure}{0.48\textwidth}
        \includegraphics[width=\linewidth]{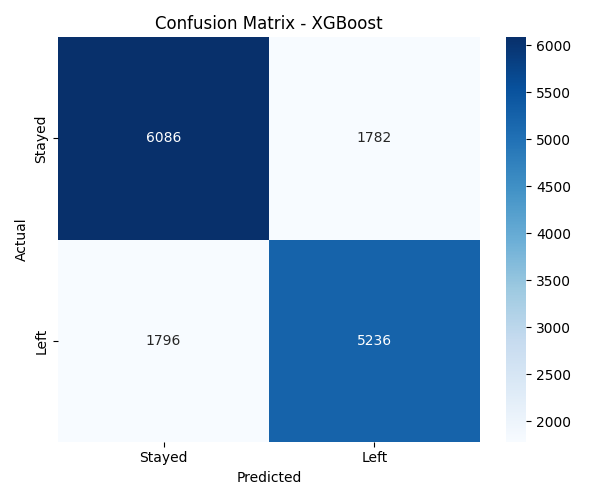}
        \caption{XGBoost}
        \label{fig:confusion_matrix_xgboost}
    \end{subfigure}
    \hfill
    \begin{subfigure}{0.48\textwidth}
        \includegraphics[width=\linewidth]{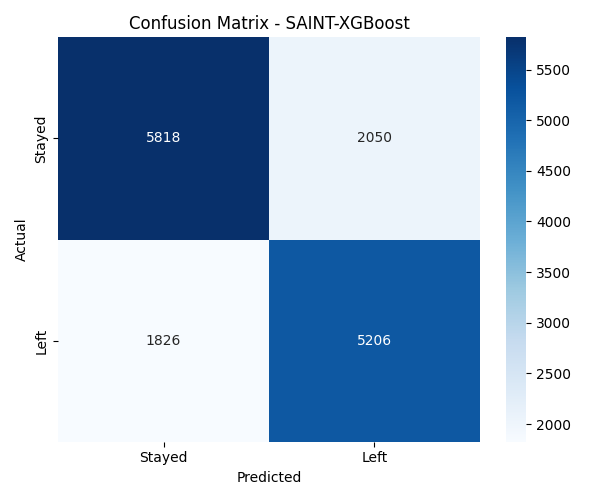}
        \caption{SAINT-XGBoost}
        \label{fig:confusion_matrix_saint_xgboost}
    \end{subfigure}

    \caption{Confusion matrices for XGBoost and SAINT-XGBoost hybrid models.}
    \label{fig:confusion_matrices_combined}
\end{figure}

\subsection{SHAP Analysis Details}
\label{sec:shap_app}
SHAP analysis provides insights into feature importance and interpretability. The following Figures present SHAP value comparisons for different models.

\begin{figure}[!ht]
    \centering

    % === Signed SHAP Values ===
    \begin{subfigure}{0.48\linewidth}
        \includegraphics[width=\linewidth]{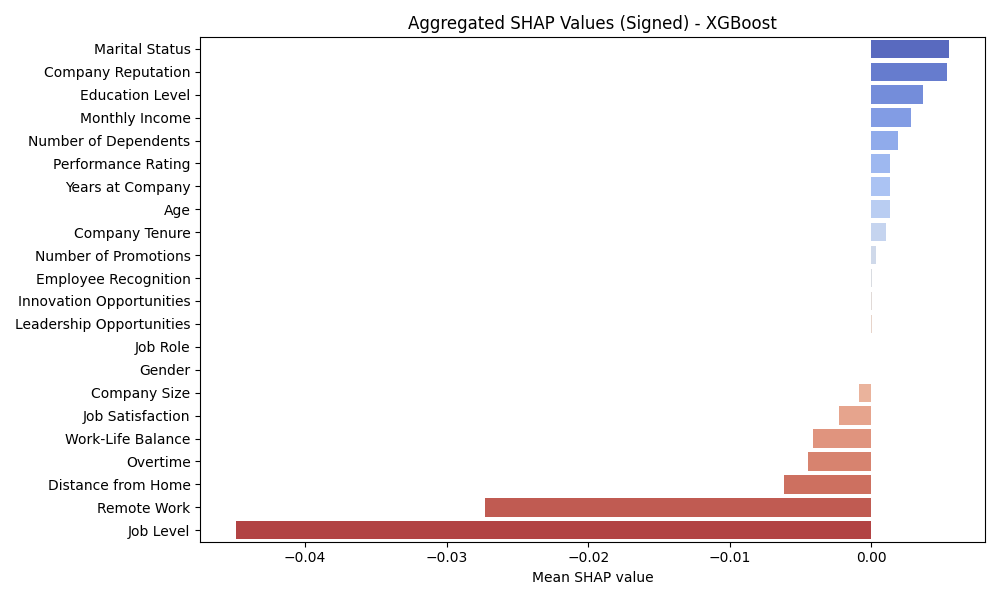}
        \caption{XGBoost}
        \label{fig:SHAP_XGBOOST_SIGN}
    \end{subfigure}
    \hfill
    \begin{subfigure}{0.48\linewidth}
        \includegraphics[width=\linewidth]{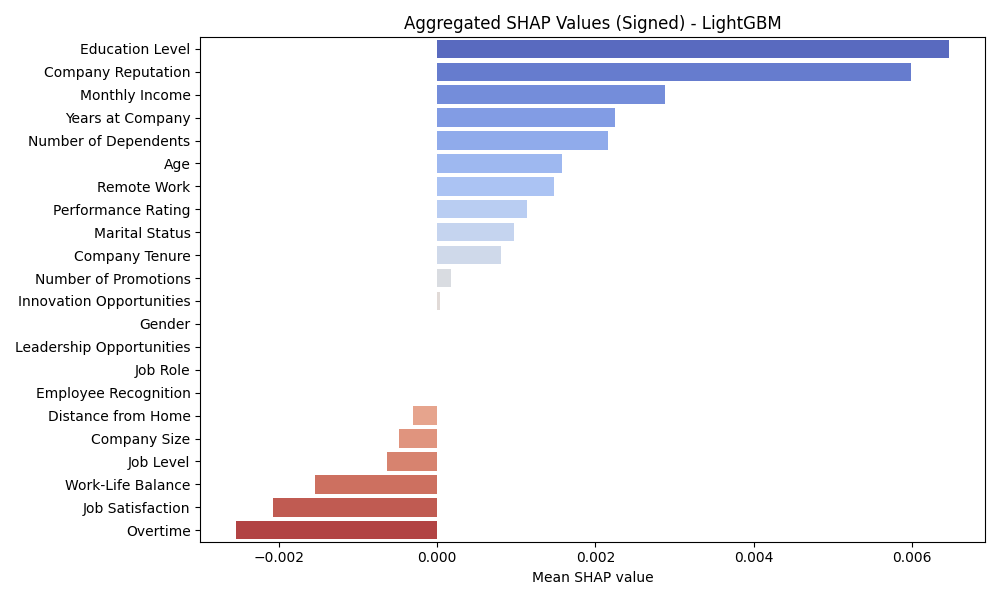}
        \caption{LightGBM}
        \label{fig:SHAP_LGBM_SIGN}
    \end{subfigure}

    \vspace{1em}

    \begin{subfigure}{0.48\linewidth}
        \includegraphics[width=\linewidth]{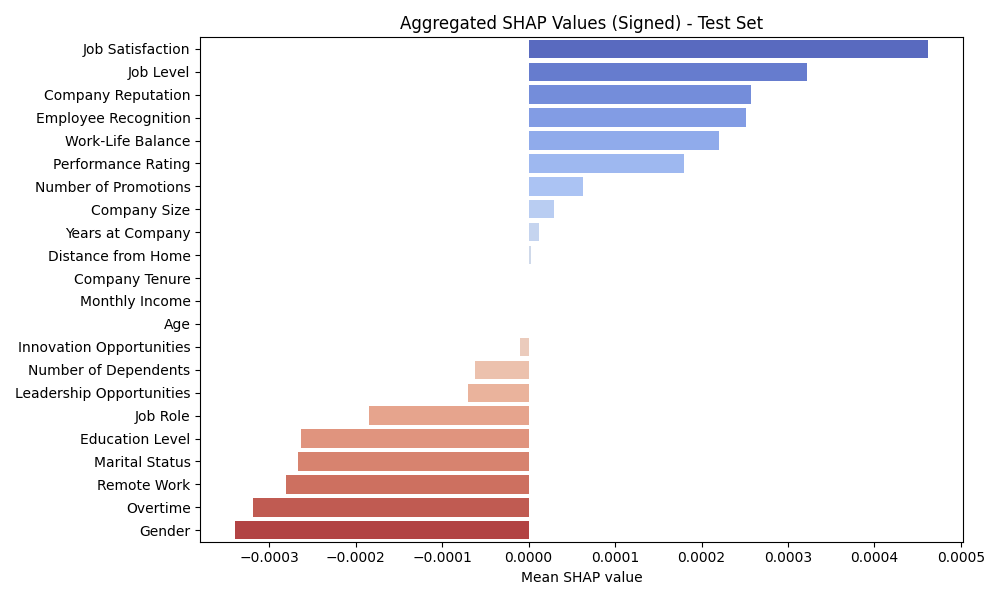}
        \caption{SAINT}
        \label{fig:SHAP_SAINT_SIGN}
    \end{subfigure}
    \hfill
    \begin{subfigure}{0.48\linewidth}
        \includegraphics[width=\linewidth]{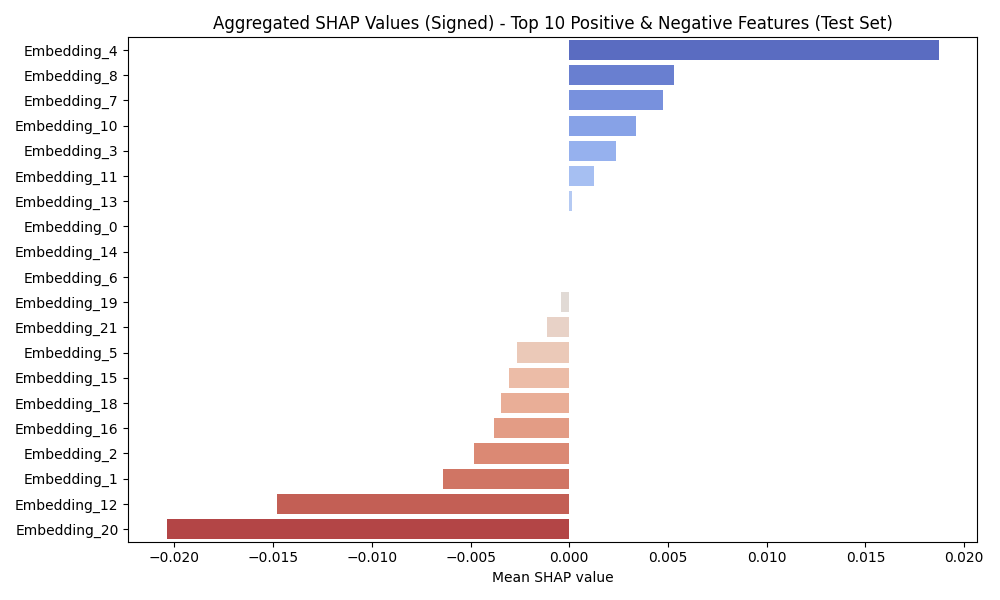}
        \caption{XGBoost-SAINT}
        \label{fig:SHAP_SAINT-XGBOOST_SIGN}
    \end{subfigure}

    \vspace{1em}

    \begin{subfigure}{0.6\linewidth}
        \includegraphics[width=\linewidth]{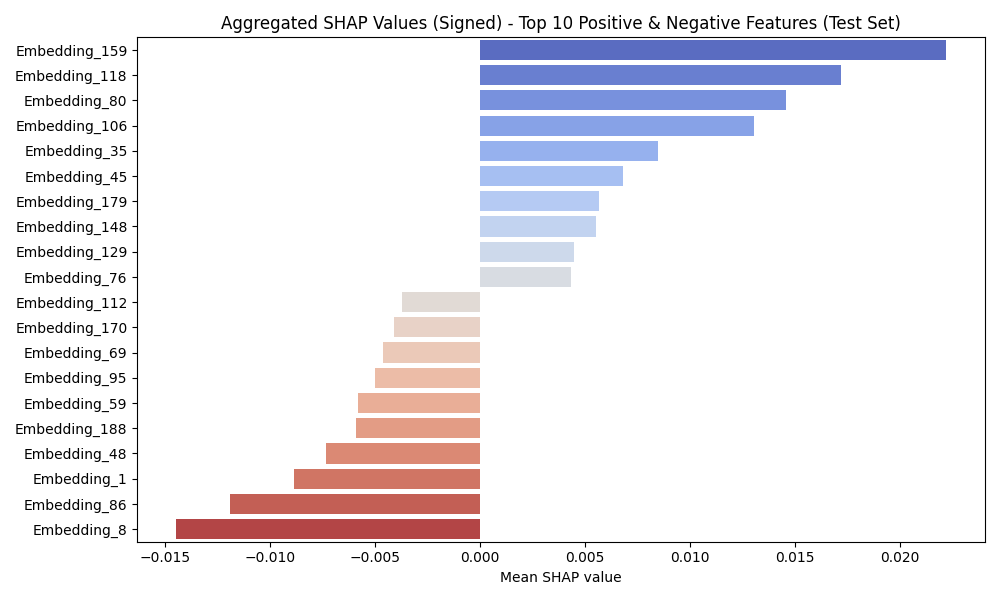}
        \caption{LightGBM-SAINT}
        \label{fig:SHAP_SAINT-LGBM_SIGN}
    \end{subfigure}

    \caption{Signed SHAP values for various models.}
    \label{fig:shap_sign_all}
\end{figure}

\clearpage

\bibliographystyle{splncs03}
\bibliography{references}

\end{document}